\documentclass[preprint,12pt,3p]{elsarticle}

\usepackage{amssymb}
\usepackage{amsmath}
\usepackage{amsfonts}
\usepackage[utf8]{inputenc}
\usepackage{algorithm}
\usepackage[noend]{algpseudocode}
\usepackage[figuresleft]{rotating}
\usepackage{graphicx} 

\usepackage{color}
\usepackage{soul}
\usepackage[bordercolor=white,backgroundcolor=gray!30,linecolor=black,colorinlistoftodos]{todonotes}

\begin{document}

\begin{frontmatter}

\title{Mixed-Integer Optimization Approach to Learning Association Rules for Unplanned ICU Transfer}


\author[label1]{Chun-An Chou}
\address[label1]{Department of Mechanical \& Industrial Engineering Department, Northeastern University, USA}

\ead{ch.chou@northeastern.edu}

\author[label1]{Qingtao Cao}
\ead{cao.q@husky.neu.edu}

\author[label5]{Shao-Jen Weng}
\address[label5]{Department of Industrial Engineering \& Enterprise Information Department, Tunghai University, Taiwan}
\ead{sjweng@thu.edu.tw}

\author[label6]{Che-Hung Tsai}
\ead{erdr2181@gmail.com}
\address[label6]{Department of Emergency Medicine, Taichung Veterans General Hospital Puli Branch, Taiwan}

\begin{abstract}
After admission to emergency department (ED), patients with critical illnesses are transferred to intensive care unit (ICU) due to unexpected clinical deterioration occurrence. Identifying such unplanned ICU transfers is urgently needed for medical physicians to achieve two-fold goals: improving critical care quality and preventing mortality. A priority task is to understand the crucial rationale behind diagnosis results of individual patients during stay in ED, which helps prepare for an early transfer to ICU. Most existing prediction studies were based on univariate analysis or multiple logistic regression to provide one-size-fit-all results. However, patient condition varying from case to case may not be accurately examined by the only judgment. In this study, we present a new decision tool using a mathematical optimization approach aiming to automatically discover rules associating diagnostic features with high-risk outcome (i.e., unplanned transfers) in different deterioration scenarios. We consider four mutually exclusive patient subgroups based on the principal reasons of ED visits: infections, cardiovascular/respiratory diseases, gastrointestinal diseases, and neurological/other diseases at a suburban teaching hospital. The analysis results demonstrate significant rules associated with unplanned transfer outcome for each subgroups and also show comparable prediction accuracy ($>$70\%) compared to state-of-the-art machine learning methods while providing easy-to-interpret symptom-outcome information. \end{abstract}

\begin{keyword}
Emergency department \sep critical care \sep unplanned ICU transfer \sep association rule \sep mixed-integer optimization 
\end{keyword}

\end{frontmatter}


\section{Introduction}
\label{sec1}

Emergency department (ED) is a core healthcare setting in hospitals and provides timely care to patients admitted with critical illness or injury. Patients with critical illness during hospitalization happen to be transferred to intensive care unit (ICU) due to unexpected clinical deterioration, such as respiration failure, multi-organ failure, cardiovascular failure, or sepsis \cite{Bapoje2011icutransfer,Shiloh2015icutransfer}. According to relevant studies, this is a critical issue in most hospitals where large amount of \textit{unplanned (or delayed) ICU transfers} after admission to ED occur their condition deteriorates adversely \cite{Simpson2005,Bapoje2011icutransfer,Tsai2014icutransfer1,Tsai2014icutransfer2,Delgado2013,Liu2012,Frost2009icutransfer}. Moreover, these unplanned ICU transfers usually lead to a higher mortality rate than those who are admitted directly to the ICU from ED. As a result, early recognition of such ICU transfers within 24-48 hours has been considered as a care quality indicator for ED practitioners, and therefore is urgently needed in order to achieve two-fold healthcare goals: improving critical care quality and preventing mortality. 

Previous retrospective research studies have shown strong statistical evidences that unplanned ICU transfers are associated with various reasons including patient condition deterioration or human errors in care \cite{Dahn2016icutransfer}. In a study \cite{Delgado2013}, there are 19\% of unplanned transfers to ICUs due to inappropriate admission triage, among which 80\% cases could be preventable. For understanding important rationale behind those associations,  a research studied a very large population of 178,315 patients, where $\sim2.4\%$ are unplanned ICU transfers within 24 hours, and found that unplanned ICU transfers are mostly associated with respiratory condition, myocardial infarction, or sepsis \cite{Dahn2016icutransfer}. In another recent work \cite{Boerma2017icutransfer}, it concluded that patients admitted to ED with hypercapnia is a high-risk group for unplanned ICU transfer, followed by patient groups with sepsis or pneumonia, in their studied database. The other research group has a focus on identifying key risk factors for unplanned ICU transfer with infections in organ systems \cite{Tsai2014icutransfer1,Tsai2014icutransfer2}. The concept of PIRO (predisposition, insult/infection, physiological response, and organ dysfunction) model for sepsis was adopted to develop their predictive system, which has a potential for the prediction of unplanned ICU transfers according to their experimental results. More formally, in critical care management, there are various well-developed scoring systems widely used for checking if patients are at risk of death. For instance, APACHE II and SAPS II are scoring tools to determine the severity level of patients using various physiological responses and clinical status as admitted in ICU \cite{knaus1985apache,le1993new}. SOFA is another similar tool that is used for tracking patients with sepsis-related organ failure in ICU \cite{vincent1998use}. These scoring systems used statistical regression analysis of selected variables converted from vital signs, lab results, and other clinical symptoms, and computed a risk probability based on large training populations \cite{le1993new,knaus1985apache,Vincent2010criticalcare}.

In the above-mentioned studies, statistical univariate analysis and multiple logistic regression analysis are used to identify key risk factors (or features) associated with high-risk patients with critical illnesses. In particular, the result of logistic regression (LR), widely used in medical diagnosis studies, is shown to be useful for patient prediction. However, it fails to differentiate patient conditions. In most cases, the same reasons (or risk factors) causing unplanned ICU transfers may not apply to all patient conditions. Instead, rule-based systems discovering distinct patterns or rules among risk factors corresponding to high-risk patients under different conditions may be a promise for adequate treatment and intervention planning in advance. To the best of our knowledge, there is not yet a study applying rule-based methods for the unplanned ICU transfer prediction. 

In this study, we formulate the unplanned ICU transfer prediction as a binary classification optimization problem. We propose a new rule-based decision tool using mixed-integer programming and association rule techniques. Using a dataset from a suburban teaching hospital, we consider four mutually exclusive patient subgroups based on the principal reason of ED visits: infections, cardiovascular/respiratory diseases, gastrointestinal diseases, and neurological/other diseases. Our contribution is to find an optimal set of association rules leading to high accuracy efficiently, subject to that association rules of key risk factors have to be most representative (i.e., maximum coverage of patients) for various deterioration occurrences during ED stay.

The organization of this paper is as follows. In Section \ref{sec2}, we describe the background of the studied problem, including data characteristics. In Section \ref{sec3}, we present the proposed method to find best association rules for prediction in detail. In Section \ref{sec4}, a real dataset collected at a ED in the teaching hospital is used for validation and verification with a comparison to other machine learning methods. In Section \ref{sec5}, We conclude the work and mention possible future work.

\section{Background and Problem Definition}\label{sec2}
\subsection{Background}
\label{sec21}
\begin{figure}[hbtp!]
\begin{center}
  \includegraphics[scale=0.35]{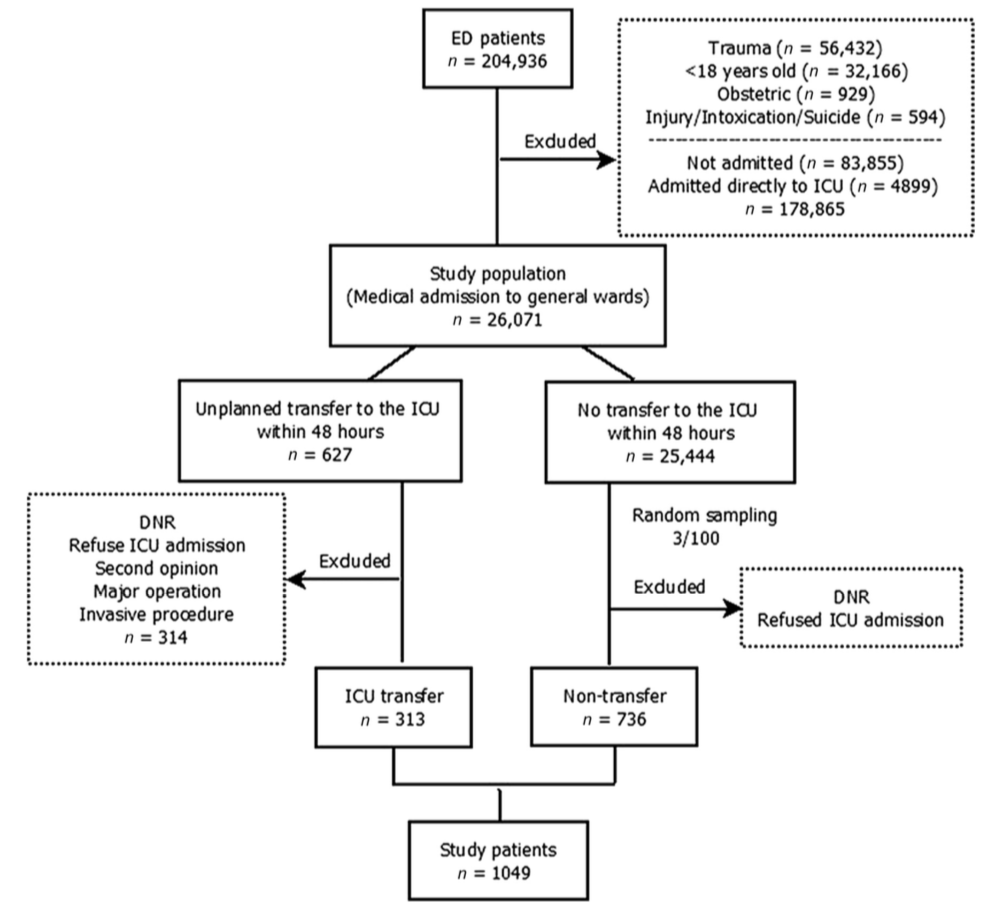}
\end{center}
\vspace{-0.2in}
  \caption{The diagram of patient selection. DNR: do not resuscitate; ED: emergency department; ICU: intensive care unit.}
\label{fig:datacollection} 
\end{figure}

In this study, a ED of a suburban teaching hospital in Taichung, Taiwan is studied. Approximate 50,000 patients annually are served in the area historically and the admission rate is $\sim25\%$, which accounts for 45\% of the inpatient population in the hospital. We retrieved the dataset between January 1, 2007, and December 31, 2010. We have a focus on patient groups with non-traumatic conditions who underwent an unplanned transfer to the ICU within 48 hours after ED admission. The control (non-transfer) group included randomly selected patients who were not transferred to the ICU within 48 hours of admission. If patients were to be admitted to a general ward but remained in the ED because of a delay or blocked access, they remained in the control group. Patients were excluded in the following conditions: (i) $<$18 years of age; (ii) admitted for injuries, intoxication, a suicide attempt, or obstetric problems; (iii) had signed ``do not resuscitate'' order; and (iv) have a critical condition but initially refused ICU admission. Patients were also excluded if they showed no clinical deterioration after admission but were transferred to the ICU for a second opinion on their potential risk. Patients who were transferred to the ICU within 48 hours for close monitoring after a major operation or invasive procedure were not enrolled in the study because these were considered expected transfers. Patients with real clinical deterioration that led to unplanned ICU transfer were the focus of this study.  In Figure \ref{fig:datacollection}, the classification of patients into the two groups: unplanned ICU transfer versus control is illustrated. We finally selected 1049 patient (736 controls and 313 unplanned transfers), and the ratio is approximately 2:1. The risk factors used in our study include basis demographics, comorbidities, organ failure history, symptoms and vital signs. The detail is described later in Section \ref{subsec41} with a summary in Table \ref{tab:datasummary}.

Two research nurses, each with at least 3 years of experience in emergency medicine or critical care, reviewed the medical records and abstracted the data on a structured data sheet. Another research assistant was responsible for data entry. One research nurse checked the data entry for accuracy. A board-certified emergency physician confirmed the quality of the data and data sheets by establishing criteria for their logical validity. The research nurses were trained on the objective of the study, the definitions of the variables, and the techniques for reviewing medical records and abstracting data. Both electronic and written medical records were reviewed. The research nurses reviewed diagnoses given during outpatient visits and hospitalizations, medications used, and results of examinations to verify the presence of certain important comorbid illnesses.

\subsection{Problem Description}
\label{sec22}
Consider a patient population presenting to an ED, emergency physicians usually make treatment decisions based on the diagnostic results, such as demographic information, clinical symptoms and signs, etc. For some cases, patients need to be transferred to ICU due to unexpected clinical deterioration. Therefore, the priority goal is to develop a decision tool that can support and determine whether or not patients are at risk to be transferred to ICU as not normally expected based on various diagnostic features (also called risk factors). Formally, this can be considered as a supervised learning problem, where the transfer state of patients is class (unplanned transfer or not) that is labeled by medical practitioners accordingly and each patient is represented by a set of diagnostic features. The objective is to discover the informative patterns (or association/decision rules) of collected diagnostic features among a large majority of patients relating to the outcome of transfer states, which then form a decision model to identify those patients who are transferred to ICU unexpectedly. For instance, an association rule is: a senior patient (\textgreater 65 years old) with chest pain history is identified as one who is most likely to be transferred to ICU. Note that the patient could be also identified by another association rule of different diagnostic features. Given these interpretable association rules, therefore, medical practitioners can manage to pay more attentions to those high-risk patients, if early recognized, as admitted from ED.

\section{Mixed-Integer Optimization Method}
\label{sec3}
In this section, we formulate the unplanned ICU transfer identification as a supervised association rule mining problem, solved by a mixed-integer optimization approach. We then describe the evaluation metrics used for validation.  

\subsection{Association Rule Formation}
\label{subsec31}
The idea of association rule was originally invented for market basket analysis and product promotion \cite{tan2005association}. Given a database of purchase records on multiple items, the goal is to discover one or more patterns (herein called association rule) showing that items $A$ and $B$ are purchased, then item $C$ is also purchased in a large portion of the records. That is, the association rule is $\{A, B\} \to \{C\}$. Equivalently, purchase records and items are related to patients and diagnostic features, respectively, in this study. 

To evaluate the goodness of association rules, the essential and widely used measure criteria are minimum \textit{support} ($\theta_{s}$) and minimum \textit{confidence} ($\theta_c$). The support $sup(ABC)$ is defined as the number of records that contain items $A$, $B$, and $C$. It is usually represented by relative support $rsup(ABC)$, the percentage of records having items $A$, $B$, and $C$ in the entire dataset. The confidence $conf(AB \to C)$ is defined as the percentage of records having item $C$ that contain items $A$ and $B$ in the entire dataset. In other words, it can be viewed as a conditional probability $Prob = sup(ABC)/sup(AB)$. As a result, a strong association rule is recognized if it is satisfied with minimum thresholds $\theta_{s}$ and $\theta_{c}$ on support and confidence, respectively. In addition, \textit{lift} is another useful measure and defined by $sup(ABC)/sup(AB)sup(C) = conf(AB \to C)/sup(C)$. If \textit{lift} = 1, there is no association between items $AB$ and $C$. If \textit{lift} is greater than 1, it shows an strong interest regarding $AB \to C$. If \textit{lift} is less than 1, it shows no interest regarding $AB \to C$.

For a relatively large dataset, there exist multiple strong association rules. Finding even better association rules among them becomes more computationally difficult as the item size increases. The computational complexity is $3^{|K|} - 2^{|K|+1} - 1$ for a dataset of $|K|$ diagnostic features. To this end, Apriori algorithm is the first and widely used efficient approach to search for informative items and then strong association rules based on the following property \cite{agrawal1994fast,Aggarwal98mininglarge}: 
\begin{equation}
    sup(AB) \leq sup(A) \text{ if } \{A\} \subset \{AB\}. \label{eq:obs} 
\end{equation}
It starts to explore with single items (e.g., $A$) and check if the support criterion is satisfied. If $sup(A) < \theta_{s}$, it is known that item $A$ is not qualified to be part of strong rules and it need not to explore multi-items that contain $A$, e.g., $AB$ or $AC$; otherwise, items $AB$ or $AC$ may be informative because $sup(AB) \geq \theta_{s}$ or $sup(AC) \geq \theta_{s}$ possibly. In such a way, the search space of possible rules are reduced largely. Consequently, strong association rules are guaranteed by meeting the confidence criterion to form a decision model. In our applied study, a patient is not identified as an ``unplanned ICU transfer'' if diagnostic features do not match any strong association rules learned from the historical patient data. 

\begin{figure}[hbtp!]
\begin{center}
  \includegraphics[scale=0.8]{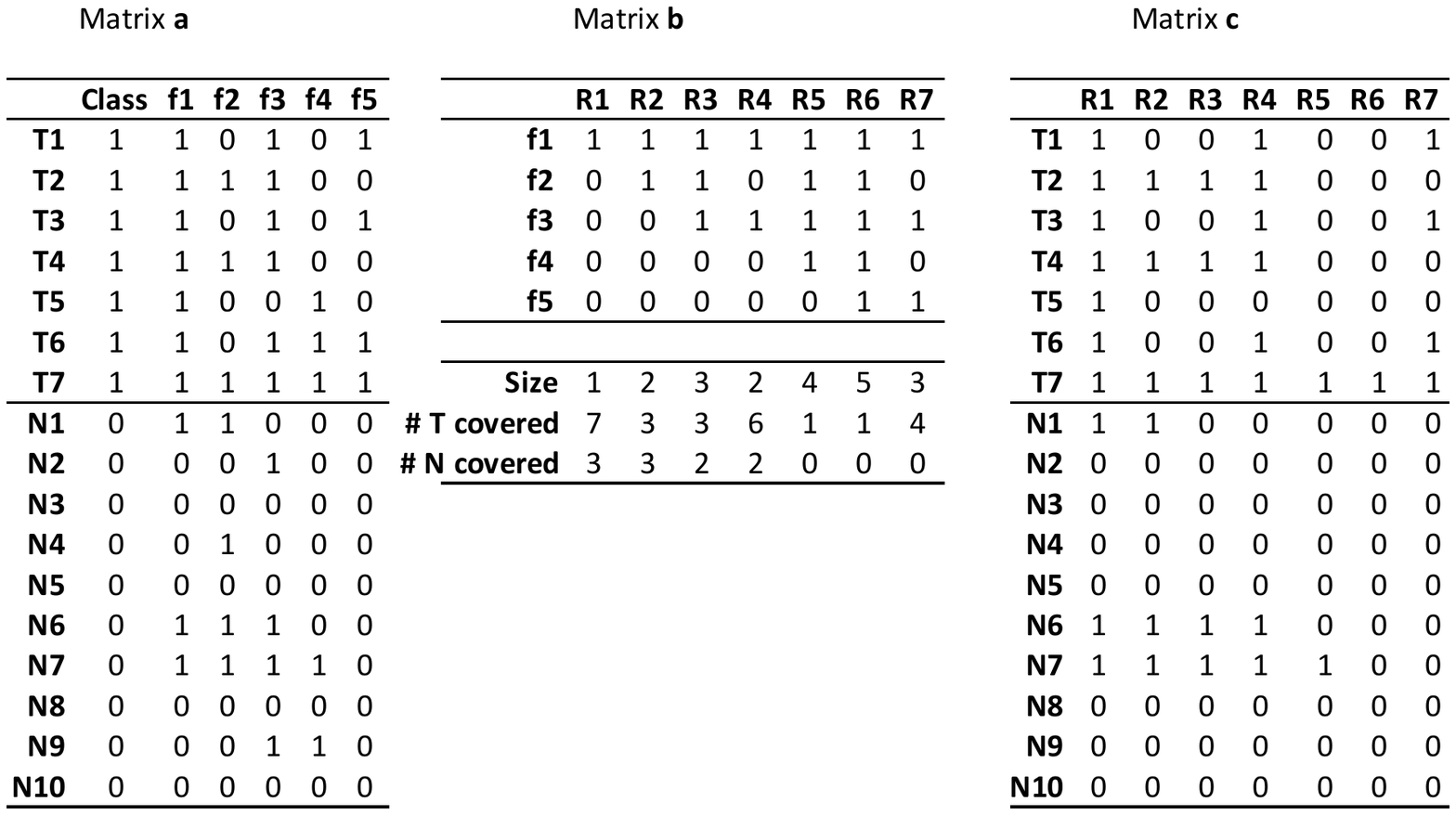}
\end{center}
\vspace{-5in}
  \caption{An illustrative example of data representation. The matrix {\bf a} on the left indicates if patients are diagnosed with features $f1-f5$ associated with ``unplanned transfer'' (Class = 1). The matrix {\bf b} in the middle represents rule information (which features are included). The matrix {\bf c} on the right indicates if patients are covered or identified by rules.}
\label{fig:toydata} 
\end{figure}

Figure \ref{fig:toydata} illustrates a toy example for association rule learning and analysis. Let us assume to have 17 patients with 5 diagnostic features, labeled in two classes: $\{Class = 1\}$ represents the target class of ``unplanned transfer'' and $\{Class = 0\}$ otherwise. Suppose $\theta_{s}$ = 5/17 and $\theta_{c}$ = 0.7. Only items $f1$ and $f3$ are eligible. Considering a feature $f1$ and class, the rule $R1: \{f1\} \to \{Class = 1\}$ is formed with support of 7/17 and confidence of 0.7, and considered as a strong association rule. By the observation in Equation (\ref{eq:obs}), multi-items such as $\{f1, f2\}$ and $\{f1, f3\}$ are possibly qualified, and the the rule $R2: \{f1, f2\} \to \{Class = 1\}$ and $R4: \{f1, f3\} \to \{Class = 1\}$ are obtained. However, $R2$ is disqualified because its support of 3 $< \theta_{s}$ = 5, and so are $R5$ and $R6$. Since $R4$ is qualified, $R7$ can be explored, but disqualified because its support of 4 $< \theta_{s}$ = 5. Among these explored association rules, one could identify $R1$ to form a decision model although it has 3 discounted ``non-transfers'' ($N1$, $N6$, and $N7$). 

\subsection{Association Rule Selection Optimization Model}
\label{subsec32}

The problem of determining (or predicting) unplanned ICU transfer is mathematically formulated as follows. By adopting parsimony assumption -- the simpler the better, the goal is to build a decision model comprised of one or more association rules that represent significant information of diagnostic features associated with unplanned ICU transfer. Each generated rule is formed of one or more diagnostic features. It is assumed that patient data ${\bf a}$ is given in a $n \times m$ binary matrix, where $a_{ij} = 1$ indicates if patient $i$ is observed with feature $j$. The patients are grouped into two classes: positive (unplanned ICU transfer) and negative (non-unplanned ICU transfer) and ${\bf a}^{+} \cup {\bf a}^{-} = {\bf a}$ and ${\bf a}^{+} \cap {\bf a}^{-} = \emptyset$. Then, it is assumed to have a set of rules presented in a $m \times p$ binary matrix ${\bf b}$, where $b_{jk} = 1$ indicates if features $j$ is included in rule $k$. Furthermore, the coverage ${\bf c} = {\bf a} \otimes {\bf b}$ is obtained in a $n \times p$ binary matrix, where $c_{ik} = 1$ indicates if patient $i$ is covered by rule $k$ because the same features appear in the patient. Similarly, positive and negative coverage are obtained: ${\bf c}^{+} \cup {\bf c}^{-} = {\bf c}$ and ${\bf c}^{+} \cap {\bf c}^{-} = \emptyset$. 

Decision variables are defined as follows: $x_i \in \{0, 1\}$ denotes a binary variable to indicate if patient $i$ can be covered or not, $y_j \in \{0, 1\}$ denotes a binary variable to indicate if feature $j$ is used in the decision model or not, and $z_k \in \{0, 1\}$ denotes a binary variable to indicate if rule $k$ is used in the decision model or not.  

The decision model is formulated as a mixed-integer program as follows:
\begin{eqnarray}
\text{(ARSOM)} \quad \min & & \alpha \sum_{j = 1}^{m} y_j + \beta \sum_{k = 1}^{p} z_k + \gamma \sum_{i \in |I|^{-} } x_i - \lambda \sum_{i = \in |I|^{+}} x_i, \label{eq1}\\
 \text{s.t.} & &\sum_{k \in K} c_{ik}^{+} z_k \geq x_i \ \forall \ i \in I^{+}, \label{eq2}\\
    & & \sum_{k \in K} c_{ik}^{-} z_k \leq M_1 x_i \ \forall \ i \in I^{-}, \label{eq3}\\
    & & \sum_{k \in K} b_{jk} z_k \leq M_2 y_j \ \forall \ j \in J, \label{eq4}\\
    & & (\sum_{i \in I^{+}} c_{ik} + \sum_{i \in I^{-}} c_{ik} - \theta_{s} |I|) z_k \geq 0, \ \forall \ k \in K \label{eq5}\\
    & & (\frac {\sum_{i \in I^{+}} c_{ik}} {\sum_{i \in I^{+}} c_{ik} + \sum_{i \in I^{-}} c_{ik}} - \theta_{c}) z_k \geq 0, \ \forall \ k \in K \label{eq6}\\
    & & (\theta_{l} - \sum_{j \in J} b_{jk}) z_k \geq 0 \ \forall \ k \in K, \label{eq7}\\
    & & x_i, y_j, z_k \ \in \{0, 1\}. \label{eq8}
\end{eqnarray}
The objective function in Equation (\ref{eq1}) is to minimize the number of features and the number of rules included in the decision model while ensuring that the rules are selected to minimize negative coverage and maximize positive coverage. $\alpha$, $\beta$, $\gamma$, and $\lambda$ are the weight parameters, which are determined by end users depending on the emphasis of the model. The constraint set in Equation (\ref{eq2}) ensures that unplanned transfer patient $i$, if covered, is covered by at least one selected rule. The constraint set in Equation (\ref{eq3}) indicates if non-transfer patient $i$ is covered by selected rules. If so, a penalty is added in the objective fuction. The constraint set in Equation (\ref{eq4}) indicates if feature $j$ is used in any selected rules. $M_1$ and $M_2$ are big numbers and set to $|K|$+1. The constraint set in Equation (\ref{eq5}) ensures that any selected rule $k$ has to be satisfied with a pre-determined minimum support threshold $\theta_{s} \in [0, 1]$. The constraint set in Equation (\ref{eq6}) ensures that any selected rule $k$ has to be satisfied with a pre-determined minimum confidence threshold $\theta_{c} \in [0, 1]$. The constraint set in Equation (\ref{eq7}) forces that the size of selected rule $k$ can not be larger than a pre-determined number $\theta_{l}$ (the maximum value is $|J|$). Equation (\ref{eq8}) is to constrain the binary decision variables $x_i$, $y_j$, and $z_k$.

\begin{figure}[hbtp!]
\begin{center}
  \includegraphics[scale=0.6]{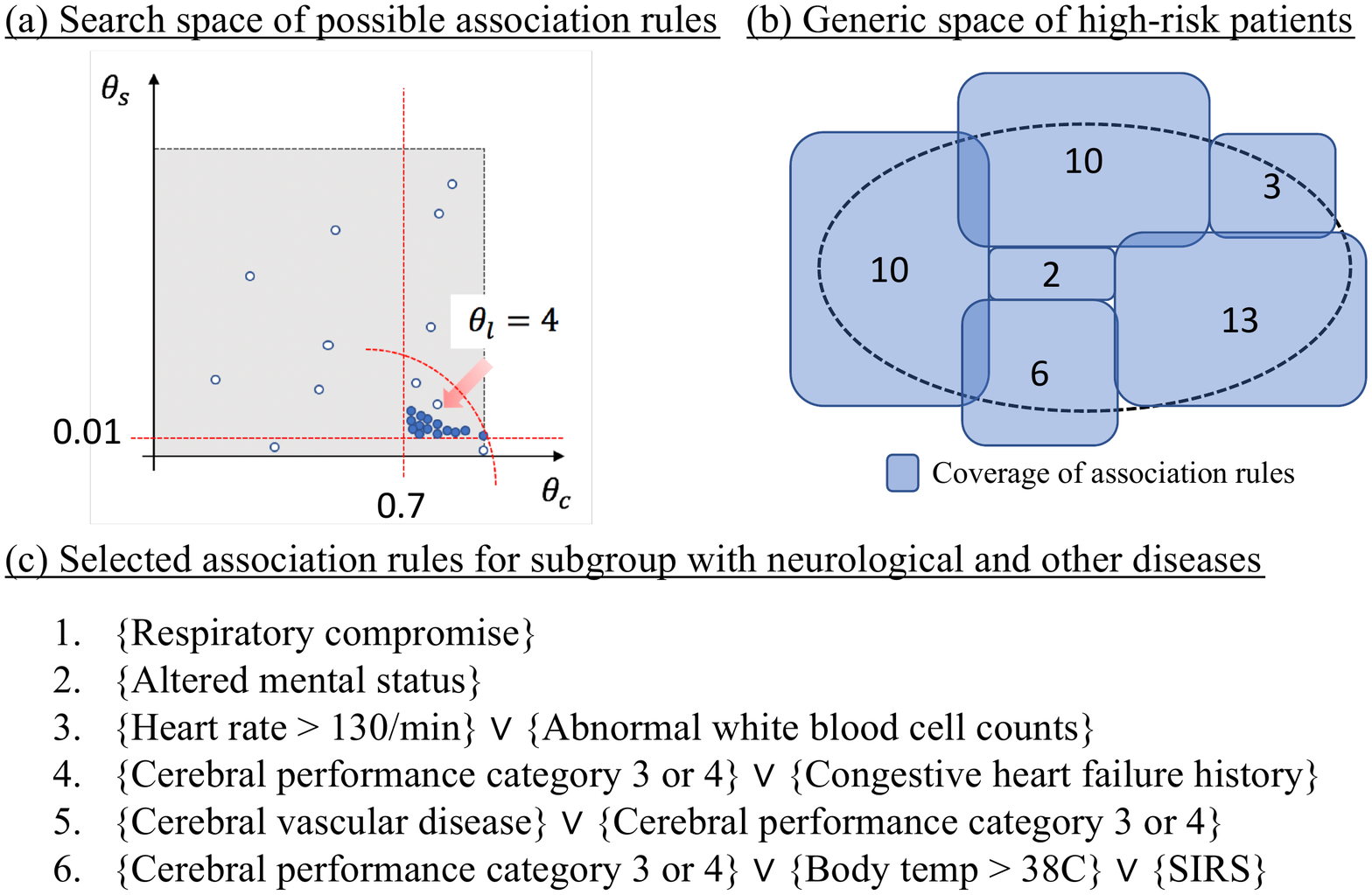}
\end{center}
\vspace{-1.5in}
  \caption{An illustration of association rules discovered by ARSOM for the subgroup 4 with neurological and other diseases. (a) The gray square area is a search space spanned by thresholds $\theta_s$ and $\theta_c$ for all possible association rules. With pre-determined thresholds $\theta_s = 0.01$, $\theta_c = 0.7$, and $\theta_l = 4$, the space is bound by red dot lines. (b) The six association rules obtained by solving ARSOM can cover all high-risk patient cases. (c) The association rules include significant features for recognizing high-risk patients.}
\label{fig:arsom} 
\end{figure}

Figure \ref{fig:arsom} illustrates the concept that the final set of strong association rules are obtained by solving the ARSOM geographically for the patient subgroup of infectious diseases. The gray square area spanned by the two parameters of support and confidence is the feasible space of all possible association rules. With pre-determined thresholds $\theta_s = 0.01$, $\theta_c = 0.7$, and $\theta_l = 4$, the space is bound by red dot lines. The thirteen blue dots represent the best set of association rules obtained by solving ARSOM to form a decision model. The detail of association analysis of diagnostic features is explained in Section \ref{sec4}. 

\subsection{Two-phase Solution Approach}
\label{subsec33}
Solving the proposed MIP model is a NP-Complete problem that is difficult to be solved exactly as the number of possible association rules increases exponentially with diagnostic feature size. We propose to decompose and solve the ARSOM model more efficiently using a two-phase heuristic. In Phase 1, we adopt the concept of association rule learning in Section \ref{subsec31} and generate strong rule candidates by Apriori algorithm which meet the three pre-determined thresholds: minimum support $\theta_{s}$, minimum confidence $\theta_{c}$, and maximum feature number $\theta_{l}$ for individual rules. Apriori algorithm is carried out using the existing package `arules' in R \cite{Hahsler2011arule}. Therefore, the constraint sets in Equations (\ref{eq5}) - (\ref{eq7}) can be dropped from the original model. The rule set $K$ becomes relatively small and contains only strong rules to select. In Phase 2, the goal of the reduced model (ARSOM-R) in Equations (\ref{eq1}) -- (\ref{eq4}) and (\ref{eq8}) is to select an optimal set of association rules that cover ``unplanned transfer'' patients at maximum and ``non-transfer'' patients at minimum. The pseudo-code for our heuristic is presented as follows. 

\begin{algorithm}
\caption{Heuristic for determining strong association rules}\label{alg1}
\begin{algorithmic}[1]
\Procedure{Run}{{$\bf a^{+}$, $\bf a^{-}$}} 
\State ${\bf R} \gets \text{Apriori}({\bf a^{+}}, \theta_s, \theta_c, \theta_l)$ \quad \Comment{{\bf R} is a set of strong association rule candidates}
\State ${\bf R_{opt}}\gets \text{ARSOM-R}({\bf R, a^{+}, a^{-}})$  \quad \Comment{{$\bf R_{opt}$} is a refined set of strong association rules}
\EndProcedure
\end{algorithmic}
\end{algorithm}

Note that in Phase 1, the strong association rules are generated for the target class of ``unplanned transfer'' only. Due to the settings of the minimum support and maximum length of rule, the generated rules may not cover all unplanned transfer patients. On one hand, those who are not covered by the generated rules are considered as outliers, which are a minority group that does not present significant information for decision making or caused by certain errors in data collection. On the other hand, this can prevent the over-fitting of our decision model in future use as model complexity increases. For instance, an association rule covering only one unplanned transfer patient is denied even though its associated confidence is one. This rule will not affect the decision outcome on new patients who may be diagnosed and identified by other strong rules with more evidences (covering a larger number of patients with similar diagnostic features). If such association rules covering minority groups are really necessary, they will be selected eventually in Phase 2. 

In the reduced model (ARSOM-R), the setting of weights $\alpha$, $\beta$, $\gamma$, and $\lambda$ in Equation (\ref{eq3}) controls the final form of decision model. If it requires to have a relatively compact decision model, $\alpha$ and $\beta$ can be set to a larger positive number to ensure smaller sets of diagnostic features and association rules, respectively, are included in the decision model. If it requires to have a more accurate decision model, $\gamma$ and $\lambda$ can be set to a larger positive number to ensure unplanned transfer patients are covered as many as possible and non-transfer patients are not covered by select rules. This is a trade-off between accuracy and model complexity, and requires a validation analysis of parameter tuning. In our study, all the parameters are set to one based on the idea that the final decision model of necessarily strong association rules cover most significant patient cases while coverage size is sufficiently large. 

\subsection{Performance Evaluation}
\label{subsec34}
To evaluate classification performance of our proposed ARSOM method, we employ the area under ROC curve (AUC) as a major evaluation metric. It is widely used for a fair judgment balancing between sensitivity and specificity, especially for problems with imbalance data between minority and majority classes. In our studied problem, true positive (TP) represents that a unplanned transferred patient (defined as positive) is identified correctly by our ARSOM classifier and false negative (FN) represents that a unplanned transferred patient is identified incorrectly. Similarly, true negative (TN) represents that a non-transferred patient (defined as negative) is identified correctly by our ARSOM classifier and false positive (FP) represents that a non-transferred patient is identified incorrectly. The sensitivity is defined to evaluate true positives and equal to TP/(TP+FN), whereas the specificity is defined to evaluate true negatives and equal to TN/(TN+FP). By varying the threshold setting, a ROC curve is formed in a sensitivity-specificity coordinate graph by connecting all dots from lower left corner to upper right corner, where a dot is represented by sensitivity and specificity. The area under this curve then represents the overall classification performance. Given a set of selected association rules in a final ARSOM, for a testing patient, we first calculate the probabilities of all selected association rules for unplanned ICU transfer. The probability $P_k$ of unplanned ICU transfer is calculated based on the confidence of an association rule $k$. A patient is classified according to the average probability ($P_a$) of all association rules that cover the patient on the same diagnostic features. The patient is identified as an unplanned transfer case if its $P_a$ is greater than a pre-set threshold $\theta_p$; otherwise. By varying threshold $\theta_p$ value applied to all testing cases, the sensitivity and specificity associated with each threshold are obtained. As a result, the ROC curve is constructed and then the AUC value is obtained. 


\section{Experimental Results}
\label{sec4}

\subsection{Data Acquisition and Preprocessing}
\label{subsec41}
Table \ref{tab:datasummary} displays a statistical summary of the data of 30 diagnostic features (risk factors) for a cohort of 1049 patients. The feature candidates include demographics (Index 1), comorbid conditions (Indices 2-6), chronic organ insufficiency (Indices 7-12), physiological responses (Indices 13-16), organ dysfunctions (Indices 17-22), and other symptoms/signs (Indices 23-30). The comorbid conditions were partly drawn from the Charlson comorbidity index \cite{Charlson1987}. Cerebral performance category (scale 1-5) is to assess neurological outcome following cardiac arrest; in this study, we considered scales 3 or 4 as a severe state (scale 5 is excluded) \cite{safar1983cpc}. Chronic organ insufficiency measurements were derived from Acute Physiology and Chronic Health Evaluation (APACHE) scores \cite{knaus1985apache}. We used the standard definition of systemic inflammatory response syndrome (SIRS) \cite{Osborn2005} for analyzing individuals' physiological responses with the thresholds of heart rate (HR) $\geq$ 130 per minute and respiratory rate (RR) $\geq$ 30 per minute. These thresholds are as the same as highest scores in the modified early warning score (MEWS) system, commonly used in Europe \cite{Subbe2001mews,Groarke2008ews}. The definitions of organ dysfunctions were drawn based on the criteria for severe sepsis \cite{Osborn2005}, except that pulmonary dysfunction was defined as an pulse oximeter oxygen saturation (SpO2) at triage \textless 90\%, a lowest SpO2 \textless 95\% with use of oxygen, or a ratio of partial pressure of oxygen to fraction of inspired oxygen (PaO2/FiO2) \textless 250 in an arterial blood gas analysis. Symptoms and signs used as calling criteria for a medical emergency team (MET) \cite{Bellomo2003} were also recorded with some modifications: respiratory compromise was defined as an RR $\geq$ 30 per minute, the presence of moderate to severe respiratory distress efforts, or an SpO2 \textless 90\% with an increased respiratory rate (RR \textgreater20 per minute). 

In this study, we grouped all patients based on the reasons for ED visits into infections, cardiovascular/respiratory diseases, gastrointestinal diseases, and neurological/other diseases, resulting in unplanned ICU transfers (UIT). All infections from any organ system were categorized as “infections” in our study, except that meningitis and central nervous system infections were categorized as neurological diseases. Intra-abdominal diseases that had developed to peritonitis and/or presented with toxic signs of infection were also categorized as infections. Because there were no predetermined criteria for emergency physicians to decide if certain tests (e.g., arterial blood gases, liver enzymes, coagulation tests, or lactate levels) would be ordered, results of tests were considered to be negative in our study if they were not ordered. 
Numerical features (e.g., age, physiological responses) are converted into a binary format using pre-defined thresholds. 

The final data for our analysis are in a binary format to indicate if diagnostic features appear or not. It is worth noting that no patients are under respiratory arrest and with heart rate \textless 40/minutes for all four categories. These two features (Indices 23 and 25) were then removed from the following implementations. 

\begin{sidewaystable}[hbtp!]
\caption{A summary of data characteristics used in this study.}
\vspace{0.1in} 
\centering
\tiny
\makebox[1 \textwidth][c]{       
\resizebox{1.3 \textwidth}{!}{   
\begin{tabular}{clrrrrrrrrrrrrrrrr}
\hline
Index & \multicolumn{1}{c}{Diagnostic feature} & \multicolumn{4}{c}{Infections (138/215)*}                                                                               & \multicolumn{4}{c}{Gastrointestinal diseases (62/179)}                                                                  & \multicolumn{4}{c}{Cardiovascular or respiratory diseases (64/82)}                                                      & \multicolumn{4}{c}{Neurological or others (49/260)}                                                                     \\
      &                                         & \multicolumn{1}{c}{UIT $\dagger$} & \multicolumn{1}{c}{percentage} & \multicolumn{1}{c}{Control $\ddagger$} & \multicolumn{1}{c}{percentage} & \multicolumn{1}{c}{UIT} & \multicolumn{1}{c}{percentage} & \multicolumn{1}{c}{Control} & \multicolumn{1}{c}{percentage} & \multicolumn{1}{c}{UIT} & \multicolumn{1}{c}{percentage} & \multicolumn{1}{c}{Control} & \multicolumn{1}{c}{percentage} & \multicolumn{1}{c}{UIT} & \multicolumn{1}{c}{percentage} & \multicolumn{1}{c}{Control} & \multicolumn{1}{c}{percentage} \\ \hline
1     & Elderly (\textgreater 65years)                    & 88                      & 63.77\%                        & 94                          & 43.72\%                        & 23                      & 37.10\%                        & 56                          & 31.28\%                        & 43                      & 67.19\%                        & 57                          & 69.51\%                        & 21                      & 42.86\%                        & 110                         & 42.31\%                        \\
2     & Diabetes                                & 67                      & 48.55\%                        & 70                          & 32.56\%                        & 24                      & 38.71\%                        & 34                          & 18.99\%                        & 25                      & 39.06\%                        & 30                          & 36.59\%                        & 13                      & 26.53\%                        & 64                          & 24.62\%                        \\
3     & Hypertension                            & 63                      & 45.65\%                        & 88                          & 40.93\%                        & 26                      & 41.94\%                        & 58                          & 32.40\%                        & 40                      & 62.50\%                        & 56                          & 68.29\%                        & 19                      & 38.78\%                        & 117                         & 45.00\%                        \\
4     & Coronary artery disease                 & 25                      & 18.12\%                        & 18                          & 8.37\%                         & 7                       & 11.29\%                        & 16                          & 8.94\%                         & 28                      & 43.75\%                        & 33                          & 40.24\%                        & 5                       & 10.20\%                        & 35                          & 13.46\%                        \\
5     & Cerebral vascular disease               & 39                      & 28.26\%                        & 39                          & 18.14\%                        & 7                       & 11.29\%                        & 11                          & 6.15\%                         & 13                      & 20.31\%                        & 14                          & 17.07\%                        & 16                      & 32.65\%                        & 50                          & 19.23\%                        \\
6     & Cerebral performance category (3 or 4)          & 47                      & 34.06\%                        & 35                          & 16.28\%                        & 3                       & 4.84\%                         & 6                           & 3.35\%                         & 13                      & 20.31\%                        & 6                           & 7.32\%                         & 14                      & 28.57\%                        & 16                          & 6.15\%                         \\
7     & Respiratory failure history             & 9                       & 6.52\%                         & 10                          & 4.65\%                         & 3                       & 4.84\%                         & 2                           & 1.12\%                         & 10                      & 15.63\%                        & 4                           & 4.88\%                         & 2                       & 4.08\%                         & 0                           & 0.00\%                         \\
8     & Congestive heart failure history        & 20                      & 14.49\%                        & 11                          & 5.12\%                         & 5                       & 8.06\%                         & 8                           & 4.47\%                         & 26                      & 40.63\%                        & 33                          & 40.24\%                        & 7                       & 14.29\%                        & 17                          & 6.54\%                         \\
9     & Liver cirrhosis history                 & 11                      & 7.97\%                         & 7                           & 3.26\%                         & 28                      & 45.16\%                        & 22                          & 12.29\%                        & 1                       & 1.56\%                         & 0                           & 0.00\%                         & 1                       & 2.04\%                         & 0                           & 0.00\%                         \\
10    & End stage renal disease                 & 11                      & 7.97\%                         & 6                           & 2.79\%                         & 4                       & 6.45\%                         & 4                           & 2.23\%                         & 6                       & 9.38\%                         & 2                           & 2.44\%                         & 3                       & 6.12\%                         & 10                          & 3.85\%                         \\
11    & Cancer                                  & 25                      & 18.12\%                        & 8                           & 3.72\%                         & 13                      & 20.97\%                        & 21                          & 11.73\%                        & 4                       & 6.25\%                         & 8                           & 9.76\%                         & 2                       & 4.08\%                         & 15                          & 5.77\%                         \\
12    & Immune compromise                       & 6                       & 4.35\%                         & 3                           & 1.40\%                         & 4                       & 6.45\%                         & 5                           & 2.79\%                         & 1                       & 1.56\%                         & 2                           & 2.44\%                         & 2                       & 4.08\%                         & 0                           & 0.00\%                         \\
13    & Body Temperature (BT) \textgreater{}$38\,^{\circ}\mathrm{C}$                   & 57                      & 41.30\%                        & 106                         & 49.30\%                        & 25                      & 40.32\%                        & 82                          & 45.81\%                        & 25                      & 39.06\%                        & 29                          & 35.37\%                        & 18                      & 36.73\%                        & 100                         & 38.46\%                        \\
14    & Heart rate \textgreater 130 / minute            & 24                      & 17.39\%                        & 18                          & 8.37\%                         & 10                      & 16.13\%                        & 6                           & 3.35\%                         & 7                       & 10.94\%                        & 5                           & 6.10\%                         & 7                       & 14.29\%                        & 9                           & 3.46\%                         \\
15    & Abnormal white blood cell counts        & 77                      & 55.80\%                        & 93                          & 43.26\%                        & 26                      & 41.94\%                        & 50                          & 27.93\%                        & 15                      & 23.44\%                        & 14                          & 17.07\%                        & 12                      & 24.49\%                        & 46                          & 17.69\%                        \\
16    & Systemic inflammatory response syndrom (SIRS)                                    & 100                     & 72.46\%                        & 104                         & 48.37\%                        & 32                      & 51.61\%                        & 39                          & 21.79\%                        & 35                      & 54.69\%                        & 27                          & 32.93\%                        & 21                      & 42.86\%                        & 37                          & 14.23\%                        \\
17    & Hypertension                             & 33                      & 23.91\%                        & 10                          & 4.65\%                         & 14                      & 22.58\%                        & 4                           & 2.23\%                         & 5                       & 7.81\%                         & 6                           & 7.32\%                         & 0                       & 0.00\%                         & 6                           & 2.31\%                         \\
18    & Respiratory compromise                  & 35                      & 25.36\%                        & 18                          & 8.37\%                         & 4                       & 6.45\%                         & 2                           & 1.12\%                         & 16                      & 25.00\%                        & 15                          & 18.29\%                        & 5                       & 10.20\%                        & 2                           & 0.77\%                         \\
19    & Renal dysfunction                       & 32                      & 23.19\%                        & 5                           & 2.33\%                         & 11                      & 17.74\%                        & 13                          & 7.26\%                         & 12                      & 18.75\%                        & 14                          & 17.07\%                        & 6                       & 12.24\%                        & 14                          & 5.38\%                         \\
20    & Liver dysfunction                       & 8                       & 5.80\%                         & 2                           & 0.93\%                         & 5                       & 8.06\%                         & 7                           & 3.91\%                         & 0                       & 0.00\%                         & 0                           & 0.00\%                         & 0                       & 0.00\%                         & 0                           & 0.00\%                         \\
21    & Hematological dysfunction               & 21                      & 15.22\%                        & 7                           & 3.26\%                         & 20                      & 32.26\%                        & 15                          & 8.38\%                         & 3                       & 4.69\%                         & 1                           & 1.22\%                         & 3                       & 6.12\%                         & 7                           & 2.69\%                         \\
22    & Metabolic dysfunction                   & 12                      & 8.70\%                         & 3                           & 1.40\%                         & 4                       & 6.45\%                         & 0                           & 0.00\%                         & 4                       & 6.25\%                         & 1                           & 1.22\%                         & 5                       & 10.20\%                        & 4                           & 1.54\%                         \\
23    & Respiratory arrest                      & 0                       & 0.00\%                         & 0                           & 0.00\%                         & 0                       & 0.00\%                         & 0                           & 0.00\%                         & 0                       & 0.00\%                         & 0                           & 0.00\%                         & 0                       & 0.00\%                         & 0                           & 0.00\%                         \\
24    & Respiratory distress                    & 35                      & 25.36\%                        & 16                          & 7.44\%                         & 5                       & 8.06\%                         & 3                           & 1.68\%                         & 15                      & 23.44\%                        & 12                          & 14.63\%                        & 6                       & 12.24\%                        & 4                           & 1.54\%                         \\
25    & Heart rate \textless 40 / minute                 & 0                       & 0.00\%                         & 0                           & 0.00\%                         & 0                       & 0.00\%                         & 0                           & 0.00\%                         & 0                       & 0.00\%                         & 0                           & 0.00\%                         & 0                       & 0.00\%                         & 0                           & 0.00\%                         \\
26    & Oliguria                                & 1                       & 0.72\%                         & 0                           & 0.00\%                         & 0                       & 0.00\%                         & 0                           & 0.00\%                         & 0                       & 0.00\%                         & 1                           & 1.22\%                         & 0                       & 0.00\%                         & 0                           & 0.00\%                         \\
27    & Altered mental status                   & 5                       & 3.62\%                         & 1                           & 0.47\%                         & 3                       & 4.84\%                         & 0                           & 0.00\%                         & 2                       & 3.13\%                         & 0                           & 0.00\%                         & 14                      & 28.57\%                        & 2                           & 0.77\%                         \\
28    & Seizure                                 & 1                       & 0.72\%                         & 0                           & 0.00\%                         & 0                       & 0.00\%                         & 0                           & 0.00\%                         & 0                       & 0.00\%                         & 0                           & 0.00\%                         & 10                      & 20.41\%                        & 10                          & 3.85\%                         \\
29    & Arrythmia                               & 0                       & 0.00\%                         & 0                           & 0.00\%                         & 0                       & 0.00\%                         & 0                           & 0.00\%                         & 3                       & 4.69\%                         & 4                           & 4.88\%                         & 0                       & 0.00\%                         & 0                           & 0.00\%                         \\
30    & Chest pain                              & 0                       & 0.00\%                         & 0                           & 0.00\%                         & 0                       & 0.00\%                         & 0                           & 0.00\%                         & 10                      & 15.63\%                        & 5                           & 6.10\%                         & 0                       & 0.00\%                         & 2                           & 0.77\%       \\
\hline
\multicolumn{18}{l}{$\dagger$ UIT stands for unplanned ICU transfer.}\\
\multicolumn{18}{l}{$\ddagger$ Control is referred to non-transferred patients.}\\
\multicolumn{18}{l}{* (\# UIT / \# Control)}
\end{tabular}
} 
} 
\label{tab:datasummary} 
\end{sidewaystable}


\subsection{Association Rule Analysis}
\label{subsec42}
In this section, we demonstrate the association rule analysis for all patients as a whole and in four subgroups separately as training sets. For each subgroup, the settings used in our ARSOM method are shown in Table \ref{tab:setting}. We use the same support thresholds $\theta_s$ = 0.01, but different conference thresholds $\theta_c$ = 0.07 for `subgroup 1' and `allgroup' versus $\theta_c$ = 0.06 for the rest since we want to include more strong association rule candidates. Note that for rule degree thresholds $\theta_{l} = 4$, it limits the maximum number of features in a rule to be 4 in for the purpose of model simplicity. We set the parameters $\alpha$, $\beta$, and $\gamma$ to be 1, and $\lambda$ to be a relatively large number in order to ensure all target patients to be covered by selected rules in the model.

In Tables \ref{tab:group1}--\ref{tab:group4} the best sets of association rules generated by our ARSOM method are presented, together with their corresponding \textit{support}, \textit{conference}, \textit{lift} and \textit{coverage}, for the four subgroups. For the infection subgroup, most significant features in comorbid conditions or chronic organ insufficiency appearing in the association rules are shown to be associated with unplanned ICU transfer. Patients with gastrointestinal disease are associated with comorbid conditions and chronic organ insufficiency, and can be observed with abnormal physiological responses. It seems obvious that unplanned ICU transfer is associated with relevant features in cardiovascular and respiratory diseases. For the subgroup neurological and other diseases, unplanned ICU transfer is could be identified by severe cerebral performance category in most cases. In Table \ref{tab:allgroup}, we also display the selected association rules as all patients are considered as a whole. It is observed that for this studied population, most pre-determined features are useful (that is, appearing in the rules with high confidence and coverage) for identifying unplanned ICU transfer. 

\begin{table}[hbtp!]
\caption{The settings in ARSOM for association rule generation for different patient subgroups.}
\vspace{0.1in} 
\centering
\scriptsize
\begin{tabular}{lccccc}
\hline
 Rule            & Subgroup 1 & Subgroup 2 & Subgroup 3 & Subgroup 4 & All  \\
                & (Infections) & (Gastrointestinal) & (Cardiovascular/respiratory) & (Neurological/others) & \\
\hline
$\theta_{s}$ & 0.01       & 0.01       & 0.01       & 0.01       & 0.01 \\
$\theta_{c}$ & 0.7        & 0.6        & 0.6        & 0.6        & 0.7  \\
$\theta_{l}$ & 4          & 4          & 4          &  4         & 4    \\
\hline
\end{tabular}
\label{tab:setting} 
\end{table}

\begin{table}[hbtp!]
\caption{Selected association rules for patient subgroup 1 with infections.}
\vspace{0.1in} 
\centering
\scriptsize  
\begin{tabular}{p{9cm}ccccc}
\hline
 Rule                            & Support & Confidence & Lift & \multicolumn{1}{p{0.5in}}{Unplanned transfer \#} & \multicolumn{1}{p{0.5in}}{Non-transfer \#} \\
\hline
\{Altered mental status\}                   & 0.01    & 0.83    & 2.13       & 5                     & 1               \\
\{Hematological dysfunction\}               & 0.06    & 0.75    & 1.92 & 21                    & 7               \\
\{Cancer\}                      & 0.07    & 0.76    & 1.94 & 25                    & 8               \\
\{Renal dysfunction\}                  & 0.09    & 0.86  & 2.21  & 32                    & 5               \\
\{Hypotension\}               & 0.09    & 0.77   & 1.96 &   33                    & 10              \\
\{Cerebral vascular disease\} $\vee$ \{End stage renal disease\}                 & 0.01    & 1.00  & 2.56  & 4                     & 0               \\
\{Cerebral vascular disease\} $\vee$ \{Respiratory failure history\}                 & 0.03    & 0.82   & 2.09 & 9                     & 2               \\
\{Coronary artery disease\} $\vee$ \{Congestive heart failure history\}                  & 0.04    & 0.93  & 2.39  & 14                    & 1               \\
\{Hypertension\} $\vee$ \{Respiratory distress\}         & 0.05    & 0.71   & 1.81 & 17                    & 7               \\
\{Diabetes\} $\vee$ \{Respiratory compromise\}             & 0.05    & 0.71   & 1.81 &  7                    & 7               \\
\{Abnormal white blood cell counts\} $\vee$ \{Respiratory compromise\}      & 0.06    & 0.74   & 1.89 & 20                    & 7               \\
\{Diabetes\} $\vee$ \{Heart rate \textgreater130\} $\vee$ \{SIRS\} & 0.02    & 0.73   & 1.86  & 8                     & 3               \\
\{Diabetes\} $\vee$ \{Coronary artery disease\} $\vee$ \{SIRS\}       & 0.04    & 0.76   & 1.96  & 13                    & 4              \\
\hline
\end{tabular}
\label{tab:group1} 
\end{table}

\begin{table}[hbtp!]
\caption{Selected association rules for patient subgroup 2 with gastrointestinal disease.}
\vspace{0.1in} 
\centering
\scriptsize
\begin{tabular}{p{8cm}ccccc}
\hline
  Rule                               & Support & Confidence & Lift & \multicolumn{1}{p{0.5in}}{Unplanned transfer \#} & \multicolumn{1}{p{0.5in}}{Non-transfer \#} \\
\hline
\{Respiratory distress\}                        & 0.02    & 0.63   & 2.43      & 5                     & 3               \\
\{Heart rate \textgreater130\}                           & 0.04    & 0.63  & 2.43      & 10                    & 6               \\
\{Hypotension\}                        & 0.06    & 0.78    & 3.02    & 14                    & 4               \\
\{End stage renal disease\} $\vee$ \{SIRS\}                    & 0.01    & 1.00   & 2.89     & 3                     & 0               \\
\{Diabetes\} $\vee$ \{Coronary artery disease\}                            & 0.02    & 0.67   & 2.59     & 4                     & 2               \\
\{Liver cirrhosis history\} $\vee$ \{Hematological dysfunction\}                    & 0.06    & 0.64   & 2.47     & 14                    & 8               \\
\{Diabetes\} $\vee$ \{Hematological dysfunction\}                    & 0.04    & 0.69   & 2.69     & 9                     & 4               \\
\{Liver cirrhosis history\} $\vee$ \{SIRS\}                     & 0.07    & 0.89  & 2.48      & 17                    & 2               \\
\{Abnormal white blood cell counts\} $\vee$ \{Liver dysfunction\} $\vee$ \{Hematological dysfunction\} & 0.01    & 0.60  & 2.33      & 3                     & 2               \\
\{Diabetes\} $\vee$ \{Hypertension\} $\vee$  \{Cerebral vascular disease\}                      & 0.02    & 0.67  & 2.59      & 4                     & 2               \\
\{Coronary artery disease\} $\vee$ \{BT \textgreater38 C\} $\vee$ \{Abnormal white blood cell counts\}         & 0.01    & 0.75  & 2.92      & 3                     & 1               \\
\{Diabetes\} $\vee$ \{Liver cirrhosis history\} $\vee$ \{Cancer\}                     & 0.01    & 0.60   & 2.33     & 3                     & 2               \\
\{Liver cirrhosis history\} $\vee$ \{BT \textgreater38 C\} $\vee$ \{Abnormal white blood cell counts\}          & 0.02    & 0.67  & 2.59      & 4                     & 2              \\
\hline
\end{tabular}
\label{tab:group2} 
\end{table}

\begin{table}[hbtp!]
\caption{Selected association rules for patient subgroup 3 with cardiovascular and respiratory diseases.}
\vspace{0.1in} 
\centering
\scriptsize
\begin{tabular}{p{9cm}ccccc}
\hline
  Rule                               & Support & Confidence & Lift & \multicolumn{1}{p{0.5in}}{Unplanned transfer \#} & \multicolumn{1}{p{0.5in}}{Non-transfer \#} \\
\hline
\{Altered mental status\}              & 0.01    & 1.00   & 2.28     & 2                     & 0               \\
\{Hematological dysfunction\}           & 0.02    & 0.75   & 1.71    & 3                     & 1               \\
\{End stage renal disease\}                  & 0.04    & 0.75  & 1.71      & 6                     & 2               \\
\{Respiratory failure history\}                  & 0.07    & 0.71    & 1.63   & 10                    & 4               \\
\{Chest pain\}             & 0.07    & 0.67   & 1.52     & 10                    & 5               \\
\{Cerebral performance category 3 or 4\}              & 0.09    & 0.68   & 1.56    & 13                    & 6               \\
\{Diabetes\} $\vee$ \{Heart rate \textgreater130\}       & 0.05    & 0.64    & 1.45    & 7                     & 4               \\
\{BT \textgreater38 C\} $\vee$  \{Abnormal white blood cell counts\} & 0.06    & 0.60  & 1.37     & 9                     & 6               \\
\{SIRS\} $\vee$ \{Respiratory compromise\} & 0.10    & 0.64   & 1.45    & 14                    & 8               \\
\{Coronary artery disease\} $\vee$ \{SIRS\}       & 0.10    & 0.65  & 1.49     & 15                    & 8               \\
\{\textgreater65years\} $\vee$ \{Coronary artery disease\} $\vee$ \{Cerebral vascular disease\}   & 0.05    & 0.62   & 1.40    & 8                     & 5               \\
\hline
\end{tabular}
\label{tab:group3} 
\end{table}

\begin{table}[hbtp!]
\caption{Selected association rules for patient subgroup 4 with neurological and other diseases.}
\vspace{0.1in} 
\centering
\scriptsize
\begin{tabular}{p{10cm}ccccc}
\hline
  Rule                               & Support & Confidence & Lift & \multicolumn{1}{p{0.5in}}{Unplanned transfer \#} & \multicolumn{1}{p{0.5in}}{Non-transfer \#} \\
\hline
\{Respiratory compromise\}                        & 0.02    & 0.71   & 4.5    & 2                     & 0               \\
\{Altered mental status\}                         & 0.05    & 0.88   & 5.52    & 3                     & 1               \\
\{Heart rate \textgreater130\} $\vee$ \{Abnormal white blood cell counts\}            & 0.02    & 0.83   & 5.26    & 6                     & 2               \\
\{Cerebral performance category 3 or 4\} $\vee$ \{Congestive heart failure history\}                   & 0.01    & 0.80  & 5.04     & 10                    & 4               \\
\{Cerebral performance category 3 or 4\} $\vee$ \{Cerebral vascular disease\}                    & 0.04    & 0.60  & 3.78     & 10                    & 5               \\
\{Cerebral performance category 3 or 4\} $\vee$ \{BT \textgreater38C\} $\vee$ \{SIRS\} & 0.02    & 0.63   & 3.94    & 13                    & 6               \\
\hline
\end{tabular}
\label{tab:group4} 
\end{table}

\begin{table}[hbtp!]
\caption{Selected association rules for the entire patient group.}
\vspace{0.1in} 
\centering
\scriptsize
\begin{tabular}{p{9cm}ccccc}
\hline
 Rule                                & Support & Confidence & Lift & \multicolumn{1}{p{0.5in}}{Unplanned transfer \#} & \multicolumn{1}{p{0.5in}}{Non-transfer \#} \\
\hline
\{Altered mental status\}                              & 0.02    & 0.89   & 2.98    & 24                    & 3               \\
\{Metabolic dysfunction\}                           & 0.02    & 0.76  & 2.54     & 25                    & 8               \\
\{Liver dysfunction\} $\vee$ \{Hematological dysfunction\}               & 0.01    & 0.73   & 2.46    & 11                    & 4               \\
\{Endstage renal disease\} $\vee$ \{SIRS\}                      & 0.02    & 0.80 & 2.68       & 16                    & 4               \\
\{Liver cirrhosis history\} $\vee$ \{Abnormal white blood cell counts\}                        & 0.02    & 0.73  & 2.45     & 19                    & 7               \\
\{Liver cirrhosis history\} $\vee$ \{SIRS\}                        & 0.02    & 0.90   & 3.00    & 26                    & 3               \\
\{Liver dysfunction\} $\vee$ \{Hematological dysfunction\}                & 0.02    & 0.89   & 2.98    & 16                    & 2               \\
\{SIRS\} $\vee$ \{Hematological dysfunction\}               & 0.03    & 0.83   & 2.79    & 30                    & 6               \\
\{Abnormal white blood cell counts\} $\vee$ \{Hypotension\}              & 0.03    & 0.76   & 2.55    & 32                    & 10              \\
\{BT \textgreater38 C\} $\vee$ \{Hypotension\}                & 0.03    & 0.75   & 2.51    & 30                    & 10              \\
\{Heart rate \textgreater130\} $\vee$ \{Respiratory distress\}              & 0.02    & 0.77  & 2.59      & 17                    & 5               \\
\{Cerebral performance category 3 or 4\} $\vee$ \{Respiratory distress\}                 & 0.03    & 0.76   & 2.54    & 28                    & 9               \\
\{Coronary artery disease\} $\vee$ \{Respiratory distress\}                    & 0.01    & 0.71   & 2.37    & 12                    & 5               \\
\{Diabetes\} $\vee$ \{Respiratory distress\}                      & 0.03    & 0.71   & 2.39    & 30                    & 12              \\
\{Abnormal white blood cell counts\} $\vee$ \{Respiratory compromise\}                 & 0.03    & 0.71    & 2.39   & 30                    & 12              \\
\{Cerebral performance category 3 or 4\} $\vee$ \{Renal dysfunction\}                  & 0.02    & 0.90   & 3.02     & 18                    & 2               \\
\{\textgreater65years\} $\vee$ \{Hypertension\} $\vee$ \{Respiratory failure history\}                  & 0.01    & 0.71   & 2.39     & 15                    & 6               \\
\{\textgreater65years\} $\vee$ \{Diabetes\} $\vee$ \{Heart rate \textgreater130\}              & 0.01    & 0.70    & 2.35    & 14                    & 6               \\
\{Diabetes\} $\vee$ \{Cancer\} $\vee$ \{Abnormal white blood cell counts\}                & 0.01    & 0.71    & 2.39    & 15                    & 6               \\
\{Diabetes\} $\vee$ \{Cancer\} $\vee$ \{SIRS\}                & 0.01    & 0.74   & 2.47     & 14                    & 5               \\
\{Hypertension\} $\vee$ \{Respiratory compromise\} $\vee$ \{Renal dysfunction\}           & 0.01    & 0.73    & 2.46    & 11                    & 4               \\
\{Coronary artery disease\} $\vee$ \{Congestive heart failure history\} $\vee$ \{SIRS\}                 & 0.02    & 0.80     & 2.68   & 24                    & 6               \\
\{\textgreater65years\} $\vee$\{Hypertension\} $\vee$\{SIRS\}$\vee$ \{Respiratory compromise\}  & 0.03    & 0.72     & 2.41   & 28                    & 11              \\
\{\textgreater65years\} $\vee$  \{Coronary artery disease\} $\vee$ \{Congestive heart failure history\} $\vee$ \{BT \textgreater38 C\}       & 0.02    & 0.71    & 2.37    & 17                    & 7               \\
\{Hypertension\} $\vee$ \{Coronary artery disease\} $\vee$ \{Abnormal white blood cell counts\} $\vee$ \{SIRS\}     & 0.01    & 0.70    & 2.35    & 14                    & 6                 \\
\hline
\end{tabular}
\label{tab:allgroup} 
\end{table}

\subsection{Performance Comparison with Other Machine learning based Prediction Methods}
\label{subsec43}
\begin{figure}[hbtp!]
\begin{center}
  \includegraphics[scale=0.7]{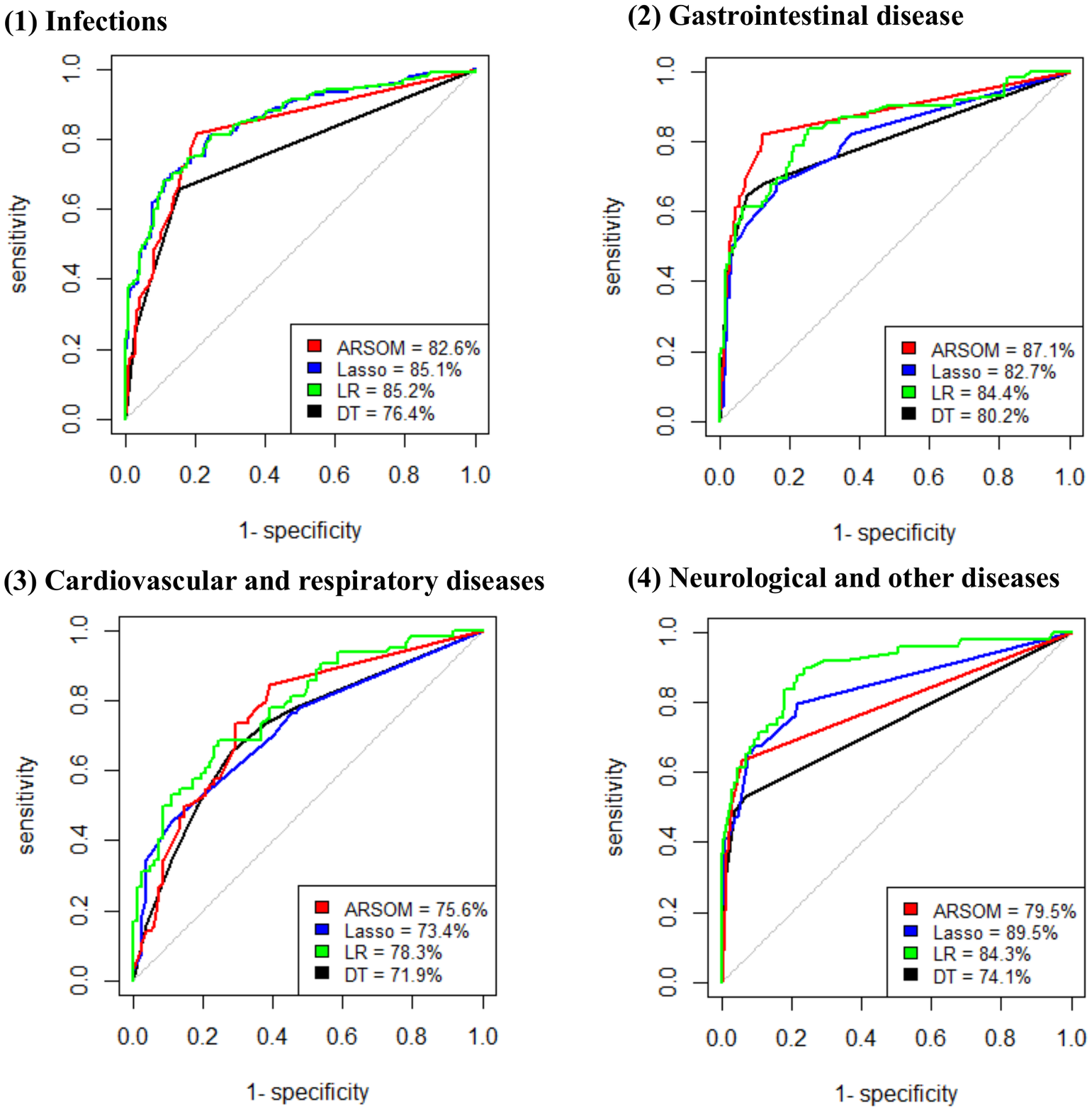}
\end{center}
\vspace{-0.2in}
  \caption{AUC Performance of ARSOM in a comparison with LR, LASSO, and DT for the four patient groups.}
\label{fig:auc} 
\end{figure}

We attempt to evaluate the classification performance of the selected association rules by our ARSOM method as compared to other machine learning methods including LR \cite{david1989} and LASSO \cite{Tibshirani1996lasso,Zou_2005}, and decision trees (DT) \cite{dt1993}. The computational results (AUC) for four patient subgroups are shown in Figure \ref{fig:auc}. The performance of our method is comparable to LR and LASSO and better than DT by around 5-8\%. 

Furthermore, we implement a out-of-sample validation using 10 times 5-fold cross validation. The computational results are presented in Table \ref{tab:comparison}, including AUC and the numbers of association rules and features included in a rule. Our method is shown to be non-inferior compared to LR and LASSO methods and slightly better than DT in terms of accuracy. For rule generation, our method generates more rules than DT. However, our rules are generated to present more individualized information while DT generates discriminative rules. Comparing to LR and LASSO, both methods can only present the discriminating power along with showing the importance of individual features, and use its one-size-fit-all result for identifying unplanned ICU transfer. 

\begin{table}[hbtp!]
\caption{Performance and model comparison with logistical regression (LR), LASSO, decision tree (DT) resulting from 10 times 5-fold cross validation.}
\vspace{0.1in} 
\centering
\footnotesize
\begin{tabular}{rrrrr}
\hline
Subgroup 1       & ARSOM        & LR & LASSO            & DT    \\
\hline
AUC            & 0.76 $\pm$ 0.01 & 0.79 $\pm$ 0.01    & 0.80 $\pm$ 0.00 & 0.72 $\pm$ 0.01 \\
Rule \#   & 13.00 $\pm$ 0.35     &           --          &         --         & 6.42 $\pm$ 0.57      \\
Feature \# & 16.30 $\pm$ 0.38     & 25.60 $\pm$ 0.01        & 21.42 $\pm$ 0.98     & 5.42 $\pm$ 0.57      \\
\hline\hline
Subgroup 2       & ARSOM        & LR & LASSO            & DT    \\
\hline
AUC            & 0.75 $\pm$ 0.03 & 0.72 $\pm$ 0.02  & 0.76 $\pm$ 0.01 & 0.74 $\pm$ 0.03 \\
Rule \#   & 8.32 $\pm$ 0.29      &        --             &        --          & 4.62 $\pm$ 0.30      \\
Feature \# & 11.36 $\pm$ 0.52     & 23.96 $\pm$ 0.08        & 8.72 $\pm$ 1.43      & 3.62 $\pm$ 0.30      \\
\hline\hline

Subgroup 3       & ARSOM        & LR & LASSO            & DT    \\
\hline
AUC            & 0.62 $\pm$ 0.03 & 0.58 $\pm$ 0.02   & 0.60 $\pm$ 0.01 & 0.52 $\pm$ 0.04 \\
Rule \#   & 8.72 $\pm$ 0.43      &        --             &        --          & 5.94 $\pm$ 0.35      \\
Feature \# & 11.22 $\pm$ 0.80     & 25.48 $\pm$ 0.14         & 9.00 $\pm$ 2.60      & 4.94 $\pm$ 0.35      \\
\hline\hline
Subgroup 4        & ARSOM        & LR & LASSO            & DT    \\
\hline
AUC            & 0.73 $\pm$ 0.02 & 0.75 $\pm$ 0.03    & 0.75 $\pm$ 0.01 & 0.70 $\pm$ 0.03 \\
Rule \#   & 5.88 $\pm$ 0.37      &        --             &         --         & 3.62 $\pm$ 0.39      \\
Feature \# & 8.90 $\pm$ 0.32      & 24.70 $\pm$ 0.11        & 8.82 $\pm$ 1.60      & 2.62 $\pm$ 0.39      \\
\hline\hline
All   & ARSOM        & LR & LASSO            &  DT    \\
\hline
AUC   & 0.73 $\pm$ 0.01               & 0.78 $\pm$ 0.00                         & 0.79 $\pm$ 0.00           & 0.72 $\pm$ 0.01                   \\
Rule \#   & 24.22 $\pm$ 0.64     &        --             &       --           & 12.54 $\pm$ 0.45     \\
Feature \# & 21.56 $\pm$ 0.45     & 27.97 $\pm$ 0.10        & 24.10 $\pm$ 0.32     & 11.54 $\pm$ 0.45 \\
\hline
\end{tabular}
\label{tab:comparison}
\end{table}

\subsection{Discussions}
\label{subsec44}
Development of prediction model decision support system is very important in many clinical situations, especially in patients at risk of unplanned ICU transfer. The traditional methods of multiple logistic regressions can only extract single risk factor, but unable to identify risk factor in combination unless defined \textit{a priori}. Our study used association rule and optimization data mining to identify risk factor of unplanned ICU transfer in ED-admitted patients, and our reduced ARSOM model was non-inferior to other machine learning methods in terms of AUC in ROC curve. 
The results of confidence and lift were also easy for clinical interpretation. If certain risk factor has a confidence of 0.75, we can say that the probability of unplanned ICU transfers under this condition is 0.75. This inference is more reliable if the support value is high. For example, in patients with infections, those patients with hematological dysfunction, history of cancer, renal dysfunction, or cardiovascular dysfunction had high confidence (\textgreater 0.75) and adequate support (\textgreater 0.06) in our study. These conditions were all compatible with diagnostic criteria of severe sepsis, which might have been neglected or ignored by emergency physician, and admitted to ordinary ward rather than direct admission to ICUs. These delayed admissions to ICU may be avoided by using clinical prediction system based on the risk factors identified by our model. In addition, effective risk factors and clinical prediction of unplanned ICU transfer is still lacking in patients with non-infections. In patients with gastrointestinal diseases, those who had history of liver cirrhosis and presented with SIRS had confidence of 0.89 and support of 0.07, which meant that these patients might have high probability of unplanned ICU transfers. Such risk group is hard to be identified by multiple logistic regressions and other traditional methods. 
The clinical inference of results of our study may be limited due to study design and small case number, but the methodology of our study could be used to develop a clinical support system, and allowed machine learning after continued input of clinical data.

\section{Conclusion and Future Work}
\label{sec5}
In this work, we presented an important problem of unplanned ICU transfer due to unexpected clinical deteriorations in critical care. We formulated it as supervised rule-based optimization problem whose main objective is to recognize high-risk patients (unplanned ICU transfer). We proposed a new rule-based decision method (ARSOM) to find the significant associations between risk factors. With the use of real data collected at the teaching hospital, we demonstrated easy-to-interpret results which is beneficial for supporting clinical decisions and non-inferior accuracy compared to other state-of-the-art machine learning methods. Because our study was retrospective, observational, and conducted at a single institution, the generalization of the presented results is considerably limited. Our conclusion would be evidently confident if a population of $>$25,000 patients admitted to medical wards (see Figure \ref{fig:datacollection}) is used to prospectively identify more patients with unplanned ICU transfer after ED admission. However, it would be almost impossible with a chart review study design. The chart reviews had flaws common to such methodologies, involving some inaccuracy and incompleteness in the measurement of vital signs and the recording of medical events as well as inconsistent criteria for ordering certain examinations and identifying abnormalities during these examinations. In the future, it is possible to include more vital signs and laboratory results in our risk evaluations. We may also adopt the concept of our proposed method and results applied to other institutions with different resources and admission policies.



\section*{Acknowledgement}
This work is support by Northeastern Faculty Startup Fund and Ministry of Science and Technology, Taiwan. We also express our special thank to the effort by Cheng-Ching General Hospital in Taichung, Taiwan for data collection, organization, explanation and sharing.







\begin{thebibliography}{10}
\expandafter\ifx\csname url\endcsname\relax
  \def\url#1{\texttt{#1}}\fi
\expandafter\ifx\csname urlprefix\endcsname\relax\def\urlprefix{URL }\fi
\expandafter\ifx\csname href\endcsname\relax
  \def\href#1#2{#2} \def\path#1{#1}\fi

\bibitem{Bapoje2011icutransfer}
S.~R. Bapoje, J.~L. Gaudiani, V.~Narayanan, R.~K. Albert, Unplanned transfers
  to a medical intensive care unit: Causes and relationship to preventable
  errors in care, Journal of Hospital Medicine 6~(2) (2011) 68--72.

\bibitem{Shiloh2015icutransfer}
A.~L. Shiloh, L.~A. Eisen, R.~H. Savel, The unplanned intensive care unit
  admission, Journal of Critical Care 30~(2) (2015) 419--420.

\bibitem{Simpson2005}
H.~K. Simpson, M.~Clancy, C.~Goldfrad, K.~Rowan, Admissions to intensive care
  units from emergency departments: a descriptive study, Emergency Medicine
  Journal 22~(6) (2005) 423--428.

\bibitem{Tsai2014icutransfer1}
J.~C.-H. Tsai, S.-J. Weng, C.-Y. Huang, D.~H.-T. Yen, H.-L. Chen, Feasibility
  of using the predisposition, insult/infection, physiological response, and
  organ dysfunction concept of sepsis to predict the risk of deterioration and
  unplanned intensive care unit transfer after emergency department admission,
  Journal of the Chinese Medical Association 77~(3) (2014) 133--141.

\bibitem{Tsai2014icutransfer2}
J.~C.-H. Tsai, C.-W. Cheng, S.-J. Weng, C.-Y. Huang, D.~H.-T. Yen, H.-L. Chen,
  Comparison of risks factors for unplanned icu transfer after ed admission in
  patients with infections and those without infections, The Scientific World
  Journal (2014) 102--929.

\bibitem{Delgado2013}
M.~K. Delgado, V.~Liu, J.~M. Pines, P.~Kipnis, M.~N. Gardner, G.~J. Escobar,
  Risk factors for unplanned transfer to intensive care within 24 hours of
  admission from the emergency department in an integrated healthcare system,
  Journal of Hospital Medicine 8~(1) (2013) 13--19.

\bibitem{Liu2012}
V.~Liu, P.~Kipnis, N.~W. Rizk, G.~J. Escobar, Adverse outcomes associated with
  delayed intensive care unit transfers in an integrated healthcare system,
  Journal of Hospital Medicine 7~(3) (2012) 224--230.

\bibitem{Frost2009icutransfer}
S.~A. Frost, E.~Alexandrou, T.~Bogdanovski, Y.~Salamonson, M.~J.Parr, K.~M.
  Hillman, Unplanned admission to intensive care after emergency
  hospitalisation: Risk factors and development of a nomogram for
  individualising risk, Resuscitation 80~(2) (2009) 224--230.

\bibitem{Dahn2016icutransfer}
C.~M. Dahn, A.~T. Manasco, A.~H. Breaud, S.~Kim, N.~Rumas, O.~Moin, P.~M.
  Mitchell, K.~P. Nelson, W.~Baker, J.~A. Feldman, A critical analysis of
  unplanned icu transfer within 48 hours from ed admission as a quality
  measure, The American Journal of Emergency Medicine 34~(8) (2016) 1505 --
  1510.

\bibitem{Boerma2017icutransfer}
L.~M. Boerma, E.~P. Reijners, R.~A. Hessels, M.~A. v~Hooft, Risk factors for
  unplanned transfer to the intensive care unit after emergency department
  admission, The American Journal of Emergency Medicine 35~(8) (2017)
  1154--1158.

\bibitem{knaus1985apache}
W.~A. Knaus, E.~A. Draper, D.~P. Wagner, J.~E. Zimmerman, {APACHE II}: a
  severity of disease classification system, Critical care medicine 13~(10)
  (1985) 818--829.

\bibitem{le1993new}
J.-R. Le~Gall, S.~Lemeshow, F.~Saulnier, A new simplified acute physiology
  score ({SAPS II}) based on a {E}uropean/{N}orth {A}merican multicenter study,
  The Journal of the American Medical Association 270~(24) (1993) 2957--2963.

\bibitem{vincent1998use}
J.-L. Vincent, A.~De~Mendon{\c{c}}a, F.~Cantraine, R.~Moreno, J.~Takala, P.~M.
  Suter, C.~L. Sprung, F.~Colardyn, S.~Blecher, Use of the {SOFA} score to
  assess the incidence of organ dysfunction/failure in intensive care units:
  results of a multicenter, prospective study, Critical care medicine 26~(11)
  (1998) 1793--1800.

\bibitem{Vincent2010criticalcare}
J.-L. Vincent, R.~Moreno, Clinical review: Scoring systems in the critically
  ill, Critical Care 14~(2) (2010) 207.

\bibitem{tan2005association}
P.-N. Tan, M.~Steinbach, V.~Kumar, Association analysis: basic concepts and
  algorithms, Introduction to data mining (2005) 327--414.

\bibitem{agrawal1994fast}
R.~Agrawal, et~al., Fast algorithms for mining association rules, in:
  Proceedings of the 20th international conference on very large data bases,
  Vol. 1215, 1994, pp. 487--499.

\bibitem{Aggarwal98mininglarge}
C.~C. Aggarwal, P.~S. Yu, Mining large itemsets for association rules, Bulletin
  of the IEEE Computer Society Technical Comittee on Data Engineering 21 (1998)
  23--31.

\bibitem{Hahsler2011arule}
M.~Hahsler, S.~Chelluboina, K.~Hornik, C.~Buchta, The arules r-package
  ecosystem: Analyzing interesting patterns from large transaction data sets,
  Journal of Machine Learning Research 12 (2011) 2021--2025.

\bibitem{Charlson1987}
M.~E.Charlson, P.~Pompei, K.~L. Ales, C.~MacKenzie, A new method of classifying
  prognostic comorbidity in longitudinal studies: Development and validation,
  Journal of Chronic Diseases 40~(5) (1987) 373--383.

\bibitem{safar1983cpc}
P.~Safar, Cerebral resuscitation after cardiac arrest: Summaries and
  suggestions, The American Journal of Emergency Medicine 1~(2) (1983)
  198--214.

\bibitem{Osborn2005}
T.~M. Osborn, H.~B. Nguyen, E.~P. Rivers, Emergency medicine and the surviving
  sepsis campaign: an international approach to managing severe sepsis and
  septic shock, Annals of Emergency Medicine 46~(3) (2005) 228--231.

\bibitem{Subbe2001mews}
C.~P. Subbe, M.~Kruger, P.~Rutherford, L.~Gemmel, Validation of a modified
  early warning score in medical admissions, Quarterly Journal of Medicine
  94~(10) (2001) 521--526.

\bibitem{Groarke2008ews}
J.~D. Groarke, J.~Gallagher, J.~Stack, A.~Aftab, C.~Dwyer, R.~McGovern,
  G.~Courtney, Use of an admission early warning score to predict patient
  morbidity and mortality and treatment success, Emergency Medicine Journal
  25~(12) (2008) 803--806.

\bibitem{Bellomo2003}
R.~Bellomo, D.~Goldsmith, S.~Uchino, J.~Buckmaster, G.~K. Hart, H.~Opdam,
  W.~Silvester, L.~Doolan, G.~Gutteridge, A prospective before-and-after trial
  of a medical emergency team, Medical Journal of Australia 179~(6) (2003)
  283--287.

\bibitem{david1989}
D.~W. Hosmer, S.~Lemeshow, Applied logistic regression, John Wiley $\&$ Sons,
  1989.

\bibitem{Tibshirani1996lasso}
R.~Tibshirani, Regression shrinkage and selection via the lasso, Journal of the
  Royal Statistical Society. Series B (Methodological) (1996) 267--288.

\bibitem{Zou_2005}
H.~Zou, T.~Hastie, Regularization and variable selection via the elastic net,
  Journal of the Royal Statistical Society: Series B (Statistical Methodology)
  67~(2) (2005) 301--320.

\bibitem{dt1993}
J.~R. Quinlan, C4.5: Programs for Machine Learning, Morgan Kaufmann, 1993.

\end{thebibliography}
\end{document}